\pdfoutput=1

\documentclass[11pt]{article}

\usepackage[final]{acl}

\usepackage{times}
\usepackage{latexsym}
\usepackage{natbib}

\usepackage[T1]{fontenc}

\usepackage[utf8]{inputenc}

\usepackage{microtype}

\usepackage{inconsolata}
\usepackage[english]{babel}

\usepackage{graphicx}

\usepackage{todonotes}
\usepackage{colortbl}
\usepackage{graphicx,multirow}

\usepackage{caption, subcaption} 
\usepackage{booktabs}

\title{An NLP Case Study on Predicting the Before and After of the Ukraine--Russia and Hamas--Israel Conflicts}

\author{Jordan Miner \\
  Hofstra University \\
  Hempstead, New York \\
  \texttt{jminer4@pride.hofstra.edu} \\\And
  John E. Ortega \\
  Hofstra University \\
  Northeastern University \\
  \texttt{john@naturallang.com} \\}

\usepackage{array}
\usepackage{multirow}
\usepackage{caption}
\usepackage{geometry}
\usepackage{booktabs}

\begin{document}
\maketitle

\begin{abstract}
We propose a method to predict toxicity and other textual attributes through the use of natural language processing (NLP) techniques for two recent events: the Ukraine--Russia and Hamas--Israel conflicts. This article provides a basis for exploration in future conflicts with hopes to mitigate risk through the analysis of social media before and after a conflict begins. Our work compiles several datasets from Twitter and Reddit for both conflicts in a 
\textit{before} and \textit{after} separation with an aim of predicting a future state of social media for avoidance. More specifically, we show that: (1) there is a noticeable difference in social media discussion leading up to and following a conflict and (2) social media discourse on platforms like Twitter and Reddit is useful in identifying future conflicts before they arise. Our results show that through the use of advanced NLP techniques (both supervised and unsupervised) toxicity and other attributes about language before and after a conflict is predictable with a low error of nearly 1.2 percent for both conflicts.
\end{abstract}

\section{Introduction}
\label{sec:intro}

In the past decade, social media has had a massive impact on how we communicate as a society in its ability to sway public opinion and shape political landscapes \cite{doi:10.1080/19331681.2017.1354243}.
In particular, the nature of the algorithms used in social networking platforms will oftentimes amplify extremist perspectives and provide users who hold these views a platform in which they can connect and share ideas \cite{Church_Schoene_Ortega_Chandrasekar_Kordoni_2023}.  It is our hypothesis that through the use of natural language processing (NLP) we could potentially help avoid social media becoming a catalyst for conflict as it has in the past.

In this study, we use NLP to examine interactions from social media on two well-known, recent conflicts: Ukraine--Russia and Hamas--Israel. We examine the role of social media in the emergence of both conflicts by gathering data from Reddit\footnote{\url{https://www.reddit.com}} and Twitter\footnote{\url{https://www.twitter.com}} and then segmenting the data into four main datasets based on date posted: (1) \textit{before} Ukraine--Russia (2) \textit{after} Ukraine--Russia (3) \textit{before} Hamas--Israel and (4) \textit{after} Hamas--Israel. 

We first reveal important insights on the segmented datasets using unsupervised techniques during development that lead to further exploration of predictive capabilities. For prediction, we use toxicity scores as a method of determining the type of language that leads up to and is used after a conflict begins based on the unsupervised results. By recognizing toxic language patterns leading up to a conflict, we can use these toxicity scores as a tool for avoidance---defined as a mechanism to prevent the escalation of a conflict by addressing or mitigating factors before they trigger or exacerbate a conflict.

Our findings show that avoidance through the use of state-of-the-art NLP techniques can be achieved on the two conflicts studied. To better illustrate our work we show that other work has not studied the more recent conflicts or used toxicity for prediction in Section \ref{sec:related}. We then illustrate the details of our dataset segmentation and methods in Section \ref{sec:methodology}. Next, in Section \ref{sec:results} and Section \ref{sec:discussion} we provide results and discussion from our experimentation. Finally, in Section \ref{sec:conclusion} we conclude with comments about achievements and next steps.

\section{Related Work}
\label{sec:related}

When used as a source of information, social media platforms' user-driven model has been known to lead to self-reinforcing polarization, a method to shape specific narratives, and act as echo chambers containing negative rhetoric to describe political or social events \cite{doi:10.1080/19331681.2017.1354243,doi:10.1080/1369118X.2016.1271900,Church_Schoene_Ortega_Chandrasekar_Kordoni_2023,10.1177/2056305120969914}. As of this paper, research in the context of both the ongoing Ukraine--Russia and Hamas--Israel have not been compared. In the past, there has been investigations about the intricacies of toxic language on social media with the Detoxify model \cite{SHETH2022312, 9806096, 9633647, 10646749}, but many of this research identified toxic content that spanned a variety of categories, rather than focusing on discussions surrounding a potential or ongoing conflict. While previous literature observes public discourse of the Ukraine--Russia conflict through the use of Latent Dirichlet Allocation (LDA) for topic modeling \cite{ASLAN2023110404, 10020274, 10234002, https://doi.org/10.1111/rsp3.12632}, many of these are used in combination with sentiment analysis only to gain an understanding of the opinions of perception of users on social media platforms like Twitter (now known as X). Additionally, LDA has also been used to observe Russian state-sponsored accounts on Twitter and their influence  2016 United States Elections \cite{10.1145/3308560.3316495}, and has been compared with alternative methods to estimate latent topics \cite{Golino_Christensen_Moulder_Kim_Boker_2021}.
Other investigations of public sentiment surrounding the Hamas--Israel conflict have taken place using sentiment analysis prior to its beginning \cite{10335101, 10.1007/978-3-031-15175-0_15}. Likewise,  \citet{Chen_He_Burghardt_Zhang_Lerman_2024} utilizes an innovative keyword extraction framework on Reddit posts created before and after the Hamas--Israel conflict, and the sentiment for a given comment was assessed using emotions like fear or sadness. Our works compares two major conflicts on a \textit{before} and \textit{after} data segmentation.  Previous research has also been carried out by \citet{doi:10.1596/1813-9450-8075} focusing on conflict prediction, but, to our knowledge, other work has not compiled the same corpora into four segmented datasets. Additionally, we provide two major aspects of prediction: topic discovery and conflict prediction for avoidance as described in Section \ref{sec:methodology}.

\section{Methodology}
\label{sec:methodology}
In this section we focus on the data collection and preparation necessary to repeat our experiments along with the model preparation for both \textit{unsupervised} discovery and \textit{supervised} prediction for avoidance. The work is made publicly available\footnote{\url{https://naturallang.com/conflict/conflict.html}} for others to consume with the aim of somehow ``sounding the alarm'' for future conflicts through social media. 
\subsection{Data Collection and Processing}
\label{sec:data_collection}
A total of four dataset were obtained to examine the role social media has in avoiding future conflicts. We again denote the datasets as the following, this time adding additional acronyms for reference purposes: (1) \textit{before} Ukraine--Russia (\textbf{URB}) (2) \textit{after} Ukraine--Russia (\textbf{URA}) (3) \textit{before} Hamas--Israel (\textbf{HIB}) and (4) \textit{after} Hamas--Israel (\textbf{HIA}).


It is noteworthy to take into account that we only processed posts in English and we feel that additional bias may have been introduced by doing so, as both conflicts took place between populations whose primary language is not English. Nonetheless, we would not want to get \textit{lost in translation} due to language differences as shown in the past \cite{van2010language}. Furthermore, the work obtained from this investigation is still helpful as it provides insight the perspectives of the international audience. In the 2014 Gaza War, social media allowed "Israel and Hamas to tailor their message to international supporters, and monitor their feedback extremely quickly" \cite{doi:10.1177/0022002716650925}. In doing so, these international supporters can then pressure their governments to choose a side in a dispute and even change the dynamics and scope.  Therefore, while international audiences might not be the directly involved, their opinions can garner political or social support in ongoing disputes that can escalate tensions into a conflict. 

URB and URA are described in the following. The first Ukraine--Russia dataset (URB) consisted of tweets posted before the conflict began with dates ranging from 31 December 2021 to 23 February 2022 \cite{Purtova_2022} that contained 835,142 documents gathered from searches including "ukraine war", "ukraine NATO", "StandWithUkraine", and "russian border ukraine" to name a few. The second Ukraine--Russia dataset (URA) was composed of tweets posted after the conflict began ranging from 24 February 2022 to 25 March 2022 \cite{bwandowando_2024}, and contained 8,268,526 documents gathered using hashtags such as "ukraineunderattack", "RussianConflict", "StopPutinNow" and "UkraineConflict" among others.

\begin{table*}[!htb]
\small
\centering
\setlength{\tabcolsep}{3pt} 
\captionsetup{font=small}
\caption{Top 5 N-grams for Each Topic by Dataset.  \\\url{https://naturallang.com/conflict/conflict.html}}
\begin{tabular}{|c|c|p{0.55\linewidth}|}
\hline
\textbf{Dataset} & \textbf{Topic} & \textbf{Top 5 Bigrams/Trigrams} \\
\hline
\multirow{9}{*}{HIB} 
 & Topic 1 & "fifa worldcup", "palestine flag", "good morning", "support palestine"  \\
 & Topic 2 & "human right", "world cup", "palestine action", "palestinian flag" \\
 & Topic 3 & "free palestine", "palestine free", "israel palestine, "israeli apartheid" \\
 & Topic 4 & "gaza strip", "palestinian people", "solidarity palestine", "day solidarity" \\
\hline
\multirow{7}{*}{HIA} 
 & Topic 1 & "sub reddit", "action performed", "bot action", "action performed automatically" \\
 & Topic 2 & "word news", "gaza strip", "hamas terrorist", "sub reddit" \\
 & Topic 3 & "west bank","middle east", "support hamas", "israeli government" \\
 & Topic 4 & "state solution", "make sense", "human shield", "sound like"\\
\hline
\multirow{9}{*}{URB} 
 & Topic 1 & "near ukraine border", "ukraine case", "troop surrounding", "nato troop"\\
 & Topic 2 & "russian star", "ukraine case", "twitter come time", "twitter com time status" \\
 & Topic 3 & "ukraine believe", "war prevent", "news euro", "twitter com time" \\
 & Topic 4 & "ukraine case", "twitter com time", "twitter com time status", "russia threat invade" \\
\hline
\multirow{7}{*}{URA} 
 & Topic 1 & "russia ukraine", "ukraine war", "ukraine russian", "ukraine ukraine" \\
 & Topic 2 & "urkaine russia", "russia war", "ukraine russia war", "war ukraine"\\
 & Topic 3 & "ukraine need", "airlift ukraine", "safe airlift", "safe airlift ukraine" \\
 & Topic 4 & "stand ukraine", "slava rain", "people ukraine", "president lensky" \\

\hline
\end{tabular}
\end{table*}

The remaining datasets (HIB and HIA) contained posts from Twitter and Reddit discussing the Hamas--Israel conflict. HIB was composed of tweets posted on Twitter before the war began with dates ranging from 1 September 2022 to 30 December 2022 \cite{Erroukrma_2022}, with a total of 24,251 documents generated from keywords mentioning "Palestine" or "Gaza." The HIA dataset consisted of posts made on Reddit from 7 October 2023 to 29 October 2023 \cite{asaniczka_2024} and contained 436,725 documents gathered from subreddits like /WorldNews and /IsraelPalestine. \\
All four of the datasets were first tokenized using the natural language toolkit\footnote{\url{https://www.nltk.org/}} (NLTK). We removed URLs, non-alphabetical characters, accents, and English stopwords. Additionaly, we tokenized the text and lemmatized using NLTK's WordNetLemmatizer\footnote{\url{https://www.nltk.org/_modules/nltk/stem/wordnet.html}}. Likewise, since Twitter is known for using hashtags, any hashtags were deconstructed into separate words using WordNinja\footnote{\url{https://github.com/keredson/wordninja}}.

Since the datasets for the Ukraine--Russia conflicts were quite large and we were limited to one GPU Tesla A100 machine with 20GB of ram, we decided to use a smaller dataset which consisted of the 174,292 URB and  1,240,279 URA documents. Size reduction was done using random sampling and stratification. Contrastingly, the HIB and HIA datasets were smaller with 20--400k documents. 

Feature vocabularies for the four datasets were first vectorized using a count vectorizer. For URB and URA, we had to limit the vocabulary to a minimum of document frequency of 5 and a maximum of 85 percent. On the other hand, the HIB and HIA datasets were set to a minimum document frequency of 5 percent and a maximum document frequency of 90. These settings resulting in a vocabulary of 5,000 terms for each corpus based on n-grams ranging from size 2 to 4. Terms for each document were combined in a term-document matrix and used in for experimentation in the \textit{unsupervised} setting that follows.



\subsection{LDA Topic Modeling}
\label{sec:methodology:sec:topic_modeling}
The unsupervised topic modeling based on LDA was used to determine whether certain documents could be grouped together based on their textual data. The optimal number of topics were obtained through experimentation to find which parameters yielded the most distinct topics and minimize any overlapping as much as possible. This yielded a total of 9 topics for the Ukraine--Russia conflict, and 7 topics for the Hamas--Israel conflict.

\subsection{Toxicity Prediction}
\label{sec:methodology:sec:toxicity_scoring}
In order to better understand how the term ``avoidance'' is deemed in this article, we present the idea of \textit{toxicity} as a prediction task. In the context of this investigation and its relevance to conflict, we define toxicity as content that fosters polarization between opposing sides, spreads distrust, and reinforces an 'us vs. them' narrative, which further encourage division and hostility. Toxic content of this nature is oftentimes used to promote the radicalization of individuals online, shape narratives about one's own group, and mobilize supporters to act \cite{d479aa59-7616-344f-918c-b0e3082b4a8f}.

It is our belief that the datasets in the two conflicts studied seem to become more toxic after a conflict had begun. This makes our task somewhat distinct from a sentiment task by digging deeper into the language, like hate speech and more, that seem to provoke and sway sentiment.

In our experiments, we used numeric toxicity values to provide an approximation between zero and one where a 0.00 toxicity score signifies not toxic at all and a score of 1.00 means extremely toxic.  We use the toxicity value because it provides a fundamental assessment of whether the text content was negative or harmful in nature so that we could examine its relevance in conflict causation.  To do this, we assigned each bag-of-word feature a toxicity score using the Detoxify\footnote{\url{https://github.com/unitaryai/detoxify}} \cite{Detoxify} library that identifies toxic content as "obscenity, threats, and identity-based hate. The toxicity scores were calculated in batches of 100, and then stored in a dictionary where each n-gram was given corresponding toxicity scores between 0.00 and 1.00.

\subsection{Linear Regression}
\label{sec:results:linear_regression_analysis:linear_regression}
We chose a \textit{supervised} linear regressor (LR) to establish a baseline toxicity prediction where URB and HIB were used to predict the toxicity scores of URA and HIA, respectively. Section \ref{sec:results} provide more insight into the original LDA results that helped show the before/after toxicity analytics. For instance, if the model predicts higher toxicity scores for social media posts after a conflict starts, toxicity and even later sentiment can be used as a mechanism of avoidance \textbf{before} a conflict hits a highly toxic point. For that reason, we attempt to predict URA and HIA toxicity with the aim of accurately predicting a future toxicity.

Independent variables for the LR model were created using document matrices similar to the \textit{unsupervised} LDA experiment. A document's toxicity score was calculated by collecting the toxicity scores of terms present in a given document, with each term associated with a calculated toxicity score described in \ref{sec:methodology:sec:toxicity_scoring}, and then calculating the average of these scores. In doing so, the LR models then used the average document scores from URB and HIB and the term-frequency matrices to predict the average toxicity scores for each document in the URA and HIA. For the entire set {URB, URA, HIB, HIA} prediction was done for individual conflicts such that URB-->URA and HIB-->HIA; we left mixing of the conflicts for future work.

To evaluate the performance of the models, the predicted toxicity scores and actual toxicity scores were compared using the mean squared error (MSE) and mean absolute error (MAE) The mean square error is the squared difference between the actual values and the predicted values, and the mean absolute examines the absolute difference; both being used to indicate how close the line of best fit is to the set of points \cite{article}. We also employed RobustScaler to scale URB and URA due to the presence of many outliers in those datasets, and MaxAbScaler for HIB and HIA as those did not contain many outliers. Both the RobustScaler and the MaxABScaler were from SciKit Learn's latest stable release, version 1.4.2.The default LR is used, which has the \textit{fit intercept} value to true. According to SciKit Learn, the default LR is: ``just plain Ordinary Least Squares (scipy.linalg.lstsq) or Non Negative Least Squares (scipy.optimize.nnls) wrapped as a predictor object''.

\subsection{BERT}
\label{sec:results:bert_model_comparison}

For comparison purposes, we compared the LR to a tranformer-basesd \cite{vaswani2017attention} model. The transformer model is a state-of-the-art model based on the BERT \cite{devlin-etal-2019-bert} architecture. This allowed us to use a pre-trained language model with the aim of transfer learning to include data from external sources along with fine-tuning on our data.

We selected the BERT model created by \citet{illinoisdatabankIDB-8882752} that had been trained on posts taken from Twitter and Youtube with the purpose of distinguishing instances of Trolling, Agression and Cyberbullying \cite{mishra-etal-2020-multilingual}. The hyperparameters used for fine-tunig/training are listed in Table \ref{tbl:bert_hyper_params}.

\begin{table}[h!]
\centering
\begin{tabular}{||c c ||} 
 \hline
 \textbf{Hyperparameter} & \textbf{Value} \\ [0.5ex] 
 \hline\hline
    fp16 & True \\ 
    epochs & 3 \\ 
    per device train batch size & 16  \\ 
    per device eval batch size & 16 \\ 
    weight decay & 0.01 \\            
    learning rate & .00002 \\ 
    save total limit & 10 \\ 
    evaluation strategy & epoch \\ [1ex] 
 \hline
\end{tabular}
\caption{Hyperparameters for training the Bert-based model using a before-->after conflict method.}
\label{tbl:bert_hyper_params}
\end{table}

We illustrate the two machine learning tasks for conflict avoidance based first on a \textit{unsupervised} technique for hypothesis approbation and then secondly with two \textit{supervised} regressors to better understand how valid our conflict avoidance hypothesis works. 


\section{Results}
\label{sec:results}
\subsection{LDA Topic Modeling}
\label{sec:results:lda_topic_modeling}

After calculating the toxicity scores of the n-grams, we wanted to inspect how the toxicity scores varied from one cluster to another. To do this, we utilized a topic-document matrix that classified documents based on their predominant topics, and a document could only be assigned to a topic so long as its highest association score was at least 80 percent. The results from this were then stored in a dictionary where each topic index was associated with a list of strongly linked documents. 

\begin{figure*}[!htb]
    \centering
        \includegraphics[width=0.80\linewidth, height=0.62\linewidth]{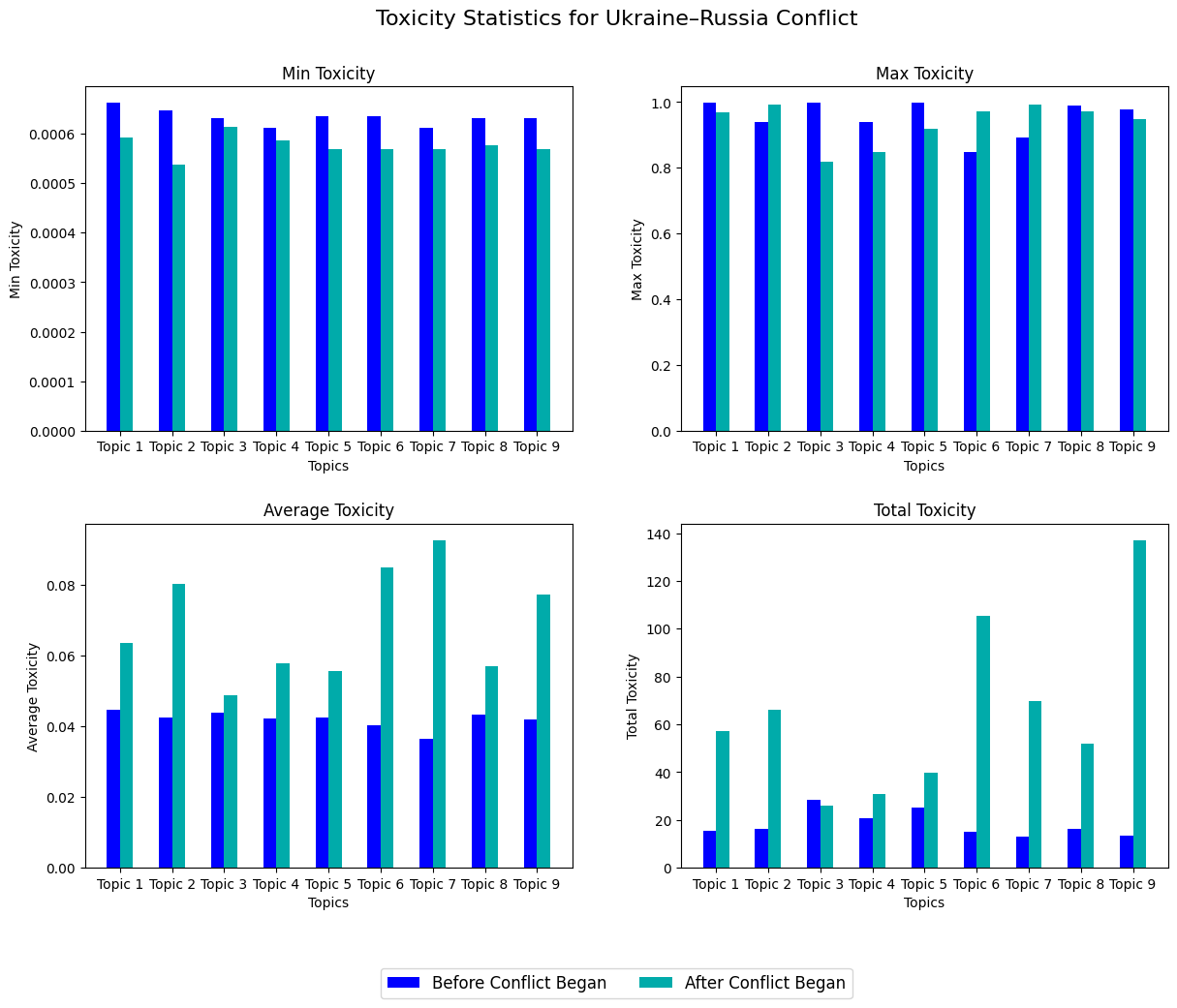} 
            \caption{Ukraine--Russia minimum, maximum, average and total toxicity of topics created with Latent Dirichlet Allocation}
            \label{sec:results:cluster_analysis:ukraine_topic_stats}
  
        \includegraphics[width=0.80\linewidth, height=0.62\linewidth]{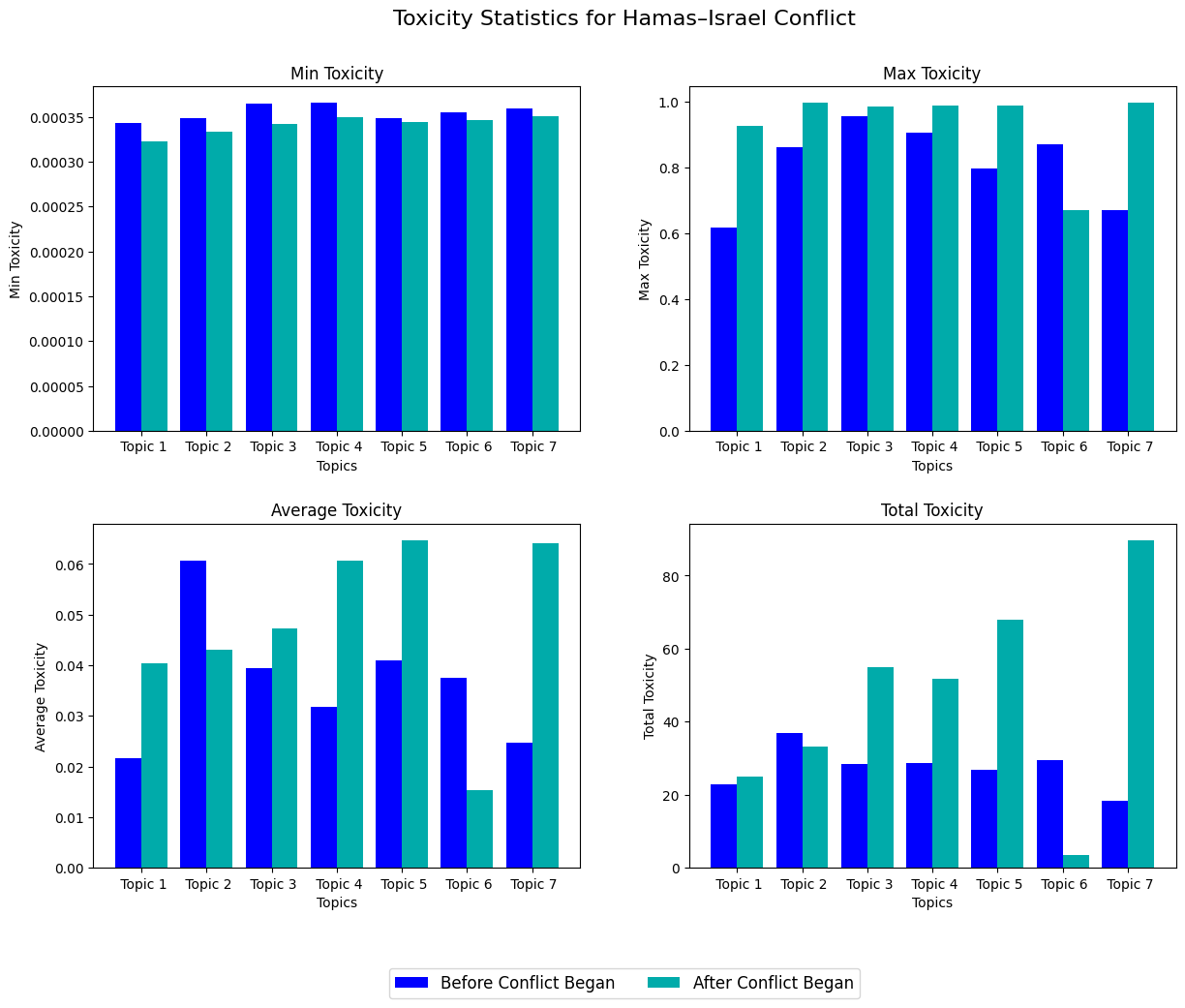}
            \caption{Hamas--Israel minimum, maximum, average and total toxicity of topics created with Latent Dirichlet Allocation}
            \label{sec:results:cluster_analysis:israel_topic_stats}  
\end{figure*}

Subsequently, by mapping the documents to the topics, and the n-grams to documents, we were then able to create a dictionary mapping topics to the n-grams, or terms, they encompassed. The resulting clusters can be visualized 
\href{https://naturallang.com/conflict/conflict.html}{online by clicking here}. Ultimately, in using this method, we  obtained the toxicity scores of each topic by extracting each term in the topic-term dictionary and matching it to the toxicity scores in the term-toxicity dictionary. The collected toxicity scores were then aggregated to compute the average, total, maximum and minimum toxicity scores for each topic as illustrated in Figures \ref{sec:results:cluster_analysis:ukraine_topic_stats} and \ref{sec:results:cluster_analysis:israel_topic_stats}.

For URB and URA, it appears that the minimum toxicity scores were mostly consistent across topics, and the minimum toxicity scores for before the conflict were slightly higher but still very close to 0.\footnote{\url{https://naturallang.com/conflict/conflict.html}} The average and total toxicity scores experienced a significant increase once the conflict began, as indicated by the higher scores for URA.The difference in toxicity were the most dramatic for Topic 6 in URB and Topic 6 in URA. Interestingly, it appeared that the URB dataset seemed to contain many extreme values since most of their maximum and minimum toxicity were higher in the URB dataset even though, on average, the URB toxicity scores were lower. Overall, it appears that, on average, most of the toxicity scores seemed to have increased upon the emergence of the war. 

 Toxicity levels between HIB and HIA also saw an increase once the conflict began. The minimum toxicity scores all appear to be the same across all topics, with the HIB toxicity scores being much lower than the HIA scores. That being said, with an exception for Topic 6 (\url{https://naturallang.com/conflict/conflict.html}), all of the maximum toxicity scores increased after the initial start of the conflict. The average and total toxicity scores, for the most part, were also much higher after the start of the conflict.

\subsection{Linear Regression and BERT}
\label{sec:results:linear_regression_analysis:linear_regression_and_bert}
In this section we compare the result of the two supervised models for accuracy according to the regression task as a manner of avoiding future conflict. We demonstrate accuracy differences for both regressors at different threshold along with the initial error in Table \ref{sec:results:model_table}.

Despite the differences in the size and content of the datasets, both models (LR and BERT) exhibit similar behaviors based on the results of the evaluation metrics. The MSE was quite small in both cases, but the lower MSE  values in the Hamas--Israel conflict suggests that the model was able to achieve a better fit to the data as it had less errors.  Similarly, for MAE, the lower the value indicates that the model also performed well with less errors, and the Hamas--Israel sets again performed better than on the Ukraine--Russia data.

\begin{figure*}[!hbt] 
    \centering 
    \begin{subfigure}[b]{0.45\linewidth}  
        \centering 
        \includegraphics[width=\textwidth, height= 0.18\textheight]{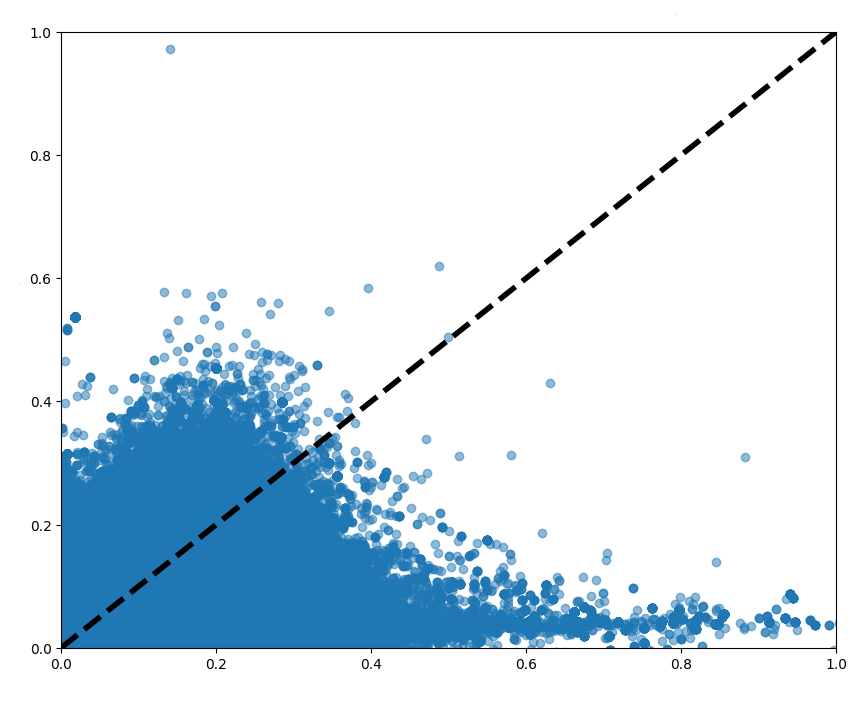} 
        \caption{Ukraine--Russia} 
        \label{fig:ukraine_lr_vis} 
    \end{subfigure} 
    \begin{subfigure}[b]{0.45\linewidth} 
        \centering 
        \includegraphics[width=\textwidth, height= 0.18\textheight]{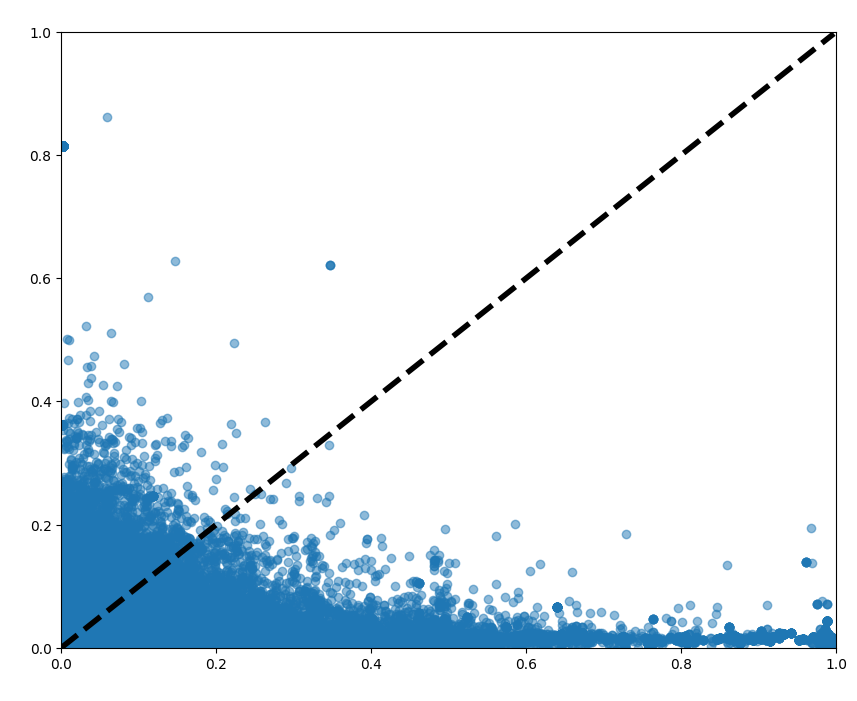} 
        \caption{Hamas--Israel} 
        \label{fig:israel_lr_vis} 
    \end{subfigure} 
    \caption{Prediction capability with Linear Regression on both conflicts using actual (x-axis) vs. predicted (y-axis) toxicity.} 
    \label{fig:scatter_comparison}  
\end{figure*}

\begin{figure*}[!hbt] 
    \centering 
    \begin{subfigure}[b]{0.45\linewidth}  
        \centering 
        \includegraphics[width=\textwidth]{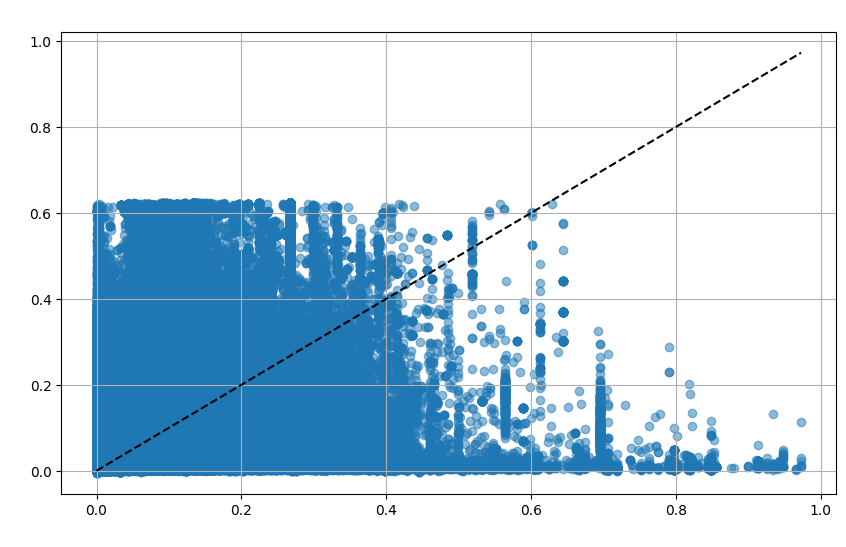} 
        \caption{Ukraine--Russia} 
        \label{fig:ukraine_hf_vis} 
    \end{subfigure} 
    \begin{subfigure}[b]{0.45\linewidth} 
        \centering 
        \includegraphics[width=\textwidth]{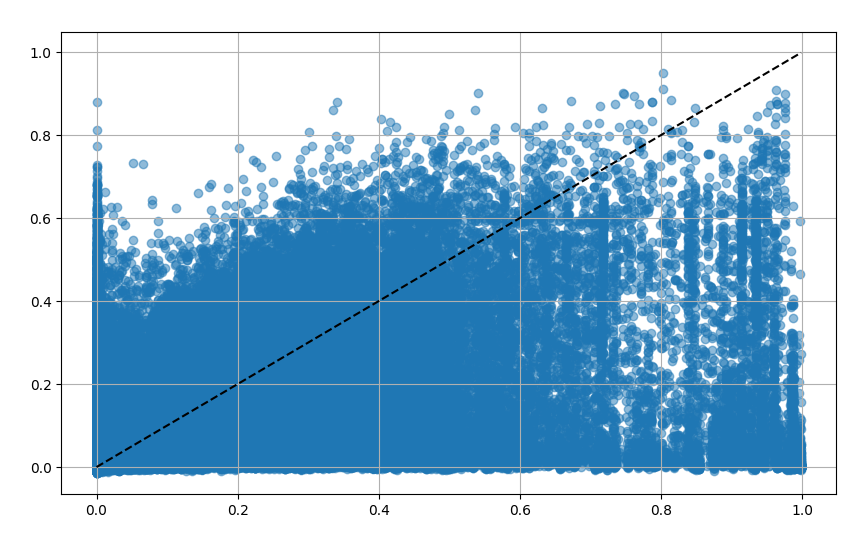} 
        \caption{Hamas--Israel} 
        \label{fig:israel_hf_vis} 
    \end{subfigure} 
    \caption{Prediction capability with BERT on both conflicts using actual (x-axis) vs. predicted (y-axis) toxicity.} 
    \label{fig:scatter_comparison_bert}  
\end{figure*}

In both scatter plots from Figures \ref{fig:scatter_comparison} (LR) and \ref{fig:scatter_comparison_bert} (BERT), the majority of the data points cluster near the bottom-left, suggesting that the majority of the actual and predicted toxicity scores were low and closer to 0.2.  For the LR model, as the actual toxicity scores increased, the Ukraine--Russia prediction scores was less likely to identify the increasing toxicity levels. This can be seen by the frequency of points that fell below the toxicity diagonal line when the actual toxicity scores were above 0.4. Thus, it can be understood that the LR model has a tendency to underestimate the magnitude of the toxicity scores, resulting in the prediction scores to be slightly lower than the actual toxicity scores.

\begin{table}
  \centering
  \begin{tabular}{lrr}
    \hline
    & \textbf{Ukr--Rus} & \textbf{Ham--Isr} \\
    \hline
    LR MSE & 0.0124 & 0.0120 \\
    LR MAE & 0.0753 & 0.0461 \\
    BERT MSE & 0.0172 & 0.0144 \\
    BERT MAE &  0.0805 & 0.0494 \\
    \hline
  \end{tabular}
  \caption{Error (as low as 1.2\%) for both the Linear Regressor and BERT models on predicting after-conflict toxicity.}
   \label{sec:results:model_table}
\end{table}

\newpage
In comparison, the LR model based on the Hamas--Israel data performed better with fewer deviations than the Ukraine-Russia model, but still struggles in identifying the highest toxicity scores. In either case, this underestimation underlines the increase in toxicity that occurred after the conflicts.

Performance of the BERT model on the two conflicts is depicted in Figure \ref{fig:scatter_comparison_bert}. For the Ukraine-Russia content, while the BERT model performs worse when measured by error alone (see Table \ref{sec:results:model_table}), its performance prediction based on the scatter plot exhibits a stronger central clustering tendency where predictions did not vary even as the actual scores changed. The better performance of our model on both datasets may speak to the relevant and informative topics extracted during the unsupervised learning portion of the investigation, as they would contribute to a better understanding of the text data. However, both models possessed a shared tendency to underestimate higher toxicity scores, as indicated by the fact that a majority of the points fell below the diagonal line. Both  models would benefit from additional fine-tuning and other methods to improve feature representation. In particular, our model may benefit from further refinement during the unsupervised portion by altering the alpha and beta parameters, or using other forms of topic modeling to improve feature quality. We save those tasks for future work.


The LR models outperform BERT in our experiments. We believe that this can be attributed to the power of small models and their objective function that has to search a smaller, more distilled space. We chose the closest pre-trained language model to our data but it could be the case that other models BERT-based or hybrid models could outperform the LR.

\subsection{Accuracy Comparison and Thresholds}
\label{sec:results:accuracy}
Various thresholds were evaluated to determine the accuracy of the model. We determined this to be the best form to measure accuracy on the level of classification alone. We believe that this would be beneficial for future use, and using one threshold over another can help balance the trade-offs between false positives and false negatives, depending on the objective of future tasks. 

\begin{figure*}[t]
    \centering
        \includegraphics[width=0.85\linewidth]{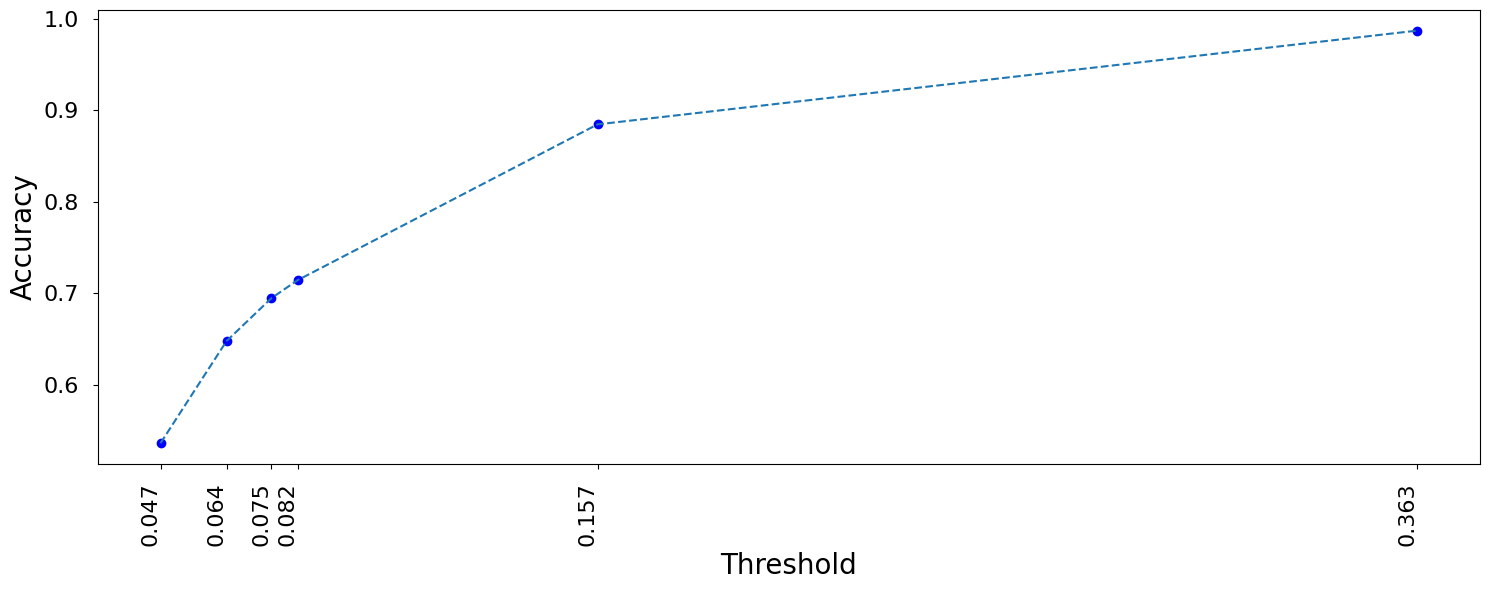} 
            \caption{Accuracy thresholds for Ukraine-Russia conflict.}
            \label{sec:results:accuracy:ukraine_russia_accuracy}

        \includegraphics[width=0.85\linewidth]{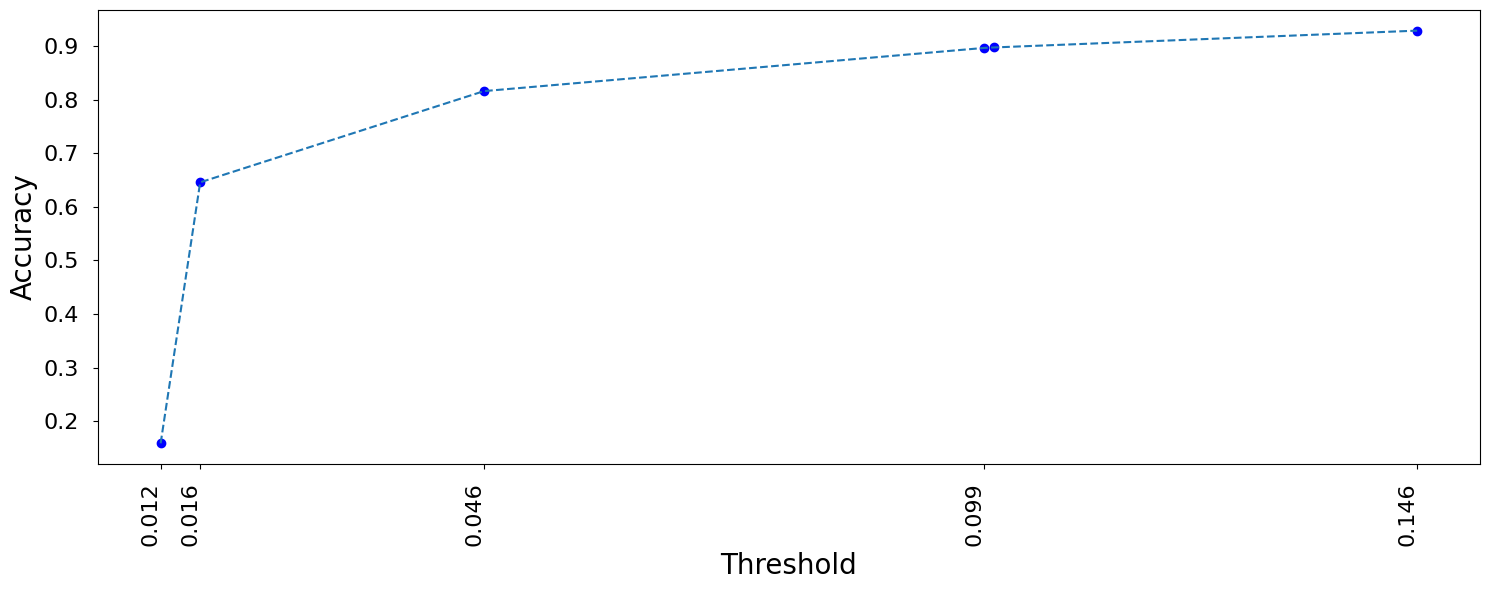}
           \caption{Accuracy thresholds for Hamas--Israel conflict.}
            \label{sec:results:accuracy:israel-hamas_accuracy}  
  
\end{figure*}

Based on the results in Figures \ref{sec:results:accuracy:ukraine_russia_accuracy} and \ref{sec:results:accuracy:israel-hamas_accuracy}, both models (LR and BERT) performed better as the threshold increased, allowing for more flexibility when it comes to determining what is considered a toxic post. For the Ukraine--Russia model, it appeared that the most optimal threshold value was the sum of the standard deviation and mean, or 0.157, and the optimal value for the Hamas--Israel model was the standard deviation of around 0.099. Hence, the optimal thresholds allow for a balance between identifying toxic posts without flagging non-toxic posts toxic or vice versa. 
These thresholds can serve as the foundation for further studies using more complex techniques to improve model reliability and accuracy. Integration of semantic analysis would also be beneficial to refine predictions that are over or under-looked using neural networks or other methods that are sensitive to complex patterns of language use. 

\section{Discussion}
\label{sec:discussion}

By incorporating LDA topic modeling, the model should have ability to detect how users' language changes during times of crisis.  We believe that the increase in total and average toxicity scores during the \textit{unsupervised} method is reflective of the overall emotion and thoughts of social media users after a conflict has begun. For instance, in the Ukraine--Russia data, the top salient terms discussed Russian troops being stationed near the eastern border and NATO's involvement to curtail war, while the post-conflict discussions focused on detailed events from the conflict and user's reactions to those events. Moreover, toxicity of certain topics experienced a noteworthy growth in comparison to others; thus indicating that certain topics were more divisive and probably elicited a stronger emotional response from user. This was seen in the case of Topic 6 (\url{https://naturallang.com/conflict/conflict.html}) in the Hamas--Israel data which contained n-grams such as ``war crime'' before the conflict, but was more heavily discussed after the conflict began.  

Furthermore, in the time leading up to the conflicts, we observed clear patterns that highlighted social media's role as an amplifier for pre-existing grievances and polarization. For the Hamas--Israel conflict, the discourse showed an increase in inflammatory content from both sides with terms like "islamic jihad" and "anti semite" to describe both sides. These terms and similar content displayed the growing distrust amid both parties, which work to feed narratives and feed existing tensions using phrases like "ethnic cleansing" and "human shield" to describe the interactions between both parties. On the other hand, the discussions prior to the onset of the Ukraine--Russia conflict also exhibited growing signs of distrust with terms like "russia invade ukraine" and "want war russia" within its rhetoric. Due to the posts being limited to English, it appeared that many of the comments painted Russia in negative light, but we would have had more conflicting perspectives had we included posts in Russian and Ukrainian.

The incorporation of LDA topics into our regression model grants it the ability to consider not only individual words, but also overarching  themes expressed, making for a more comprehensive approach that enhances prediction accuracy. Our model's ability to accurately predict post-conflict toxicity scores from pre-conflict toxicity scores indicate that these social media discussions contain early indications of unrest. While an increase in polarizing content and grievances surrounding a particular topic may not always lead directly to escalations, this toxic content can exacerbate tensions and make the conflict more likely. This would mean that governments and NGOs can monitor situations and topics that that signal growing unrest or societal division, and be immediately alerted when signs of escalation becomes prevalent and its associated toxicity levels reach a predefined critical point that could signify an increased likelihood of a conflict taking place. Furthermore, policymakers and social media platforms can use this predictive tool to gain an understanding into the language and behavioral patterns and language being used in response to events like elections or international crises in real-time.This would give policymakers and authorities the ability to address the grievances, trigger diplomatic interventions, and other peace-keeping measures to mitigate the ongoing tensions. 

Further optimizations can be implemented both by governments and social media platforms to prevent a conflict from arising. This could mean that the model would be helpful in thematically and geographically pinpointing where online toxicity is concentrated. For instance, if a certain region or group engages in more toxic content, the model would be able to pinpoint these areas as potential conflict zones and communities experiencing growing unrest. Social media platforms can also work to provide warning signs to users and strengthen moderation efforts in stances where a conflict is likely to occur. This could manifest in posts with high predicted toxicity scores to be flagged for review by human moderators and hidden from public view. In fact, developers may be able to tailor these interventions for individual users based on their predicted toxicity score in the form of warnings or temporary suspensions. To maintain engagement, these platforms can instead implement methods by elevating the voices of experts in a specified topic to prevent the spread of misinformation, and discourage instances of hate speech with customized interventions before it can incite violence.

\section{Conclusion}
\label{sec:conclusion}
Through the implementation of unsupervised and supervised machine-learning models, we have explored and observed how social media interactions can predict the escalation of two major conflicts. Particularly in times of crisis, negative sentiments and extremist perspectives are amplified on platforms like Twitter and Reddit. Furthermore, the limited regulation and addictive nature of these algorithms make these platforms effective tools for spreading misinformation and swaying public opinion, making them a catalyst for conflicts. With further fine-tuning and optimization, our models should have the ability to effectively predict a rise in toxicity in user interactions in real time. Such improvements will help policymakers and social media platforms obtain a better grasp of the dynamics of social media leading up to and during a conflict. What is more, they can help in developing frameworks to mitigate hostility with customized content moderation, and even predict disputes before they can occur. In particular, prior knowledge of a conflict is pivotal as it gives policymakers or other leaders the opportunity to act appropriately, and even formulate the proper measures to maintain peace and prevent the escalation of violence. 

\section{Limitations}
\label{sec:limitations}
Our results show that an uneven distribution of toxicity scores can heavily impact performance. In our experiments, this was most evident in the low MSE and MAE values for the Ukraine--Russia models despite being unable to properly distinguish the toxicity scores higher than 0.4, and would only be the case if the majority of data points were predicted to be low and their actual toxicity scores were low. This led to the Ukraine--Russia models having a tendency to bias towards lower toxicity scores in its predictions. Likewise, while the Hamas--Israel models performed better overall, they also experienced difficulty in the upper range, which further points to the too few high-toxicity examples. It is likely that all of the models' performances would improve if trained on a balanced training set to allow the models to effectively capture the nuances in the relationship between the text and their toxicity scores. 

Additionally, the settings for minimum document frequency in the vectorization process may have negatively impacted the toxicity scores. The point of setting the minimum document frequency is to ensure that the vectorizer would extract important terms that will serve as predictors by filtering out excess noise. On the other hand, not sufficiently adjusting the maximum document frequency may have allowed overly frequent terms to dominate the feature set, further obscuring meaningful analysis. This was definitely the case as some of the terms in the topics were unrelated with the Ukraine-Russia content containing mentions of cryptocurrency and the Hamas--Israel content containing references to actions related to the platform. Correcting these thresholds could help eliminate this noise and enhance the model's ability to perform a more nuanced toxicity analysis.

Another potential reason for the models' performance was the variation in the number of samples in the training and testing sets.  Since we were using pre-existing datasets, we were limited to what was available in only that dataset. The post-war datasets were significantly larger than the pre-war datasets, and likely may have compromised the models' ability to generalize based on their training set. This size mismatch likely affected the models' performance.
\section{Acknowledgements}
\label{sec:acknowledgements}
We would like to express our most sincerest gratitude for equipment and space contributions from Hofstra University. Additionally, we highly acknowledge Northeastern University for continued funding and support with other logistics. Lastly, website hosting and servers were provided by the owner and operators of \url{https://www.naturallang.com}.

\section*{Ethical Considerations}
We have not used any human subjects for our experimentation. Nor do we express any opinion on the two conflicts studied. 
\newpage

\bibliography{main}

\end{document}